\pdfoutput=1
\documentclass[11pt]{article}

\usepackage{emnlp2021}

\usepackage{times}
\usepackage{latexsym}

\usepackage[T1]{fontenc}
\usepackage[utf8]{inputenc}

\usepackage{microtype}
\usepackage{hyperref}
\usepackage{booktabs}
\usepackage{multirow}
\usepackage{graphicx}
\usepackage{adjustbox}
\usepackage{amsmath}
\usepackage{amsthm}
\usepackage{amssymb}
\usepackage{amsfonts}
\usepackage{algorithm}
\usepackage{mathtools}
\usepackage{algorithmic}
\usepackage{comment}
\usepackage{subcaption}
\usepackage{enumitem}

\newtheorem{definition}{Definition}
\newtheorem{theorem}{Theorem}[section]

\newtheorem{innercustomthm}{Theorem}
\newenvironment{customthm}[1]
  {\renewcommand\theinnercustomthm{#1}\innercustomthm}
  {\endinnercustomthm}
\newtheorem{innercustomlem}{Lemma}
\newenvironment{customlem}[1]
  {\renewcommand\theinnercustomlem{#1}\innercustomlem}
  {\endinnercustomlem}

\def\bbC{\mathbb{C}}
\def\bbR{\mathbb{R}}
\def\bbQ{\mathbb{Q}}

\def\calM{\mathcal{M}}
\def\calV{\mathcal{V}}

\def\calKG{\mathcal{KG}}
\def\calE{\mathcal{E}}
\def\calR{\mathcal{R}}

\def\bfI{\mathbf{I}}

\def\bfi{\mathbf{i}}
\def\bfj{\mathbf{j}}
\def\bfk{\mathbf{k}}

\def\Appdix[#1]{\textcolor{red}{Appendix #1}}

\title{BiQUE: Biquaternionic Embeddings of Knowledge Graphs}

\author{Jia Guo \and Stanley Kok \\
        Department of Information Systems and Analytics,
        School of Computing  \\
        National University of Singapore\\
          \texttt{guojia@u.nus.edu, skok@comp.nus.edu.sg}}

\begin{document}
\maketitle

\begin{abstract}
Knowledge graph embeddings (KGEs) compactly encode multi-relational knowledge graphs (KGs).
Existing KGE models rely on geometric operations to model relational patterns.
Euclidean (circular) rotation is useful for modeling patterns such as symmetry, but cannot represent hierarchical semantics. In contrast, hyperbolic models are effective at modeling hierarchical relations, but do not perform as well on patterns on which circular rotation excels.
It is crucial for KGE models to unify multiple geometric transformations so as to fully cover the multifarious relations in KGs. To do so, we propose BiQUE, a novel model that employs 
{\it biquaternions} to integrate multiple geometric transformations, viz., scaling, translation, Euclidean rotation, and hyperbolic rotation. BiQUE makes the best trade-offs among geometric operators during training, picking the best one (or their best combination) for each relation. 
Experiments on five datasets show BiQUE's effectiveness.%
\end{abstract}

\section{Introduction}
\label{sec:intro}

Knowledge graphs (KGs) provide an efficient way to represent real-world entities and their intricate connections in the form of (\textit{head}, \textit{relation}, \textit{tail}) triples. Each \textit{head}/\textit{tail} entity corresponds to a node in a KG, and each \textit{relation} represents a directed edge between them. Imbued with rich factual knowledge, KGs have demonstrated their effectiveness in a wide range of downstream applications~\cite{WangZWZCZZC18,saxenaACL20}.
Although problems due to incompleteness and noise continue to plague KGs, those issues have been ameliorated by knowledge graph embeddings (KGEs) that project entities and relations into low-dimensional dense vectors.

\begin{figure}[t]
\centering
\includegraphics[height=3.8cm, width=0.49\textwidth]{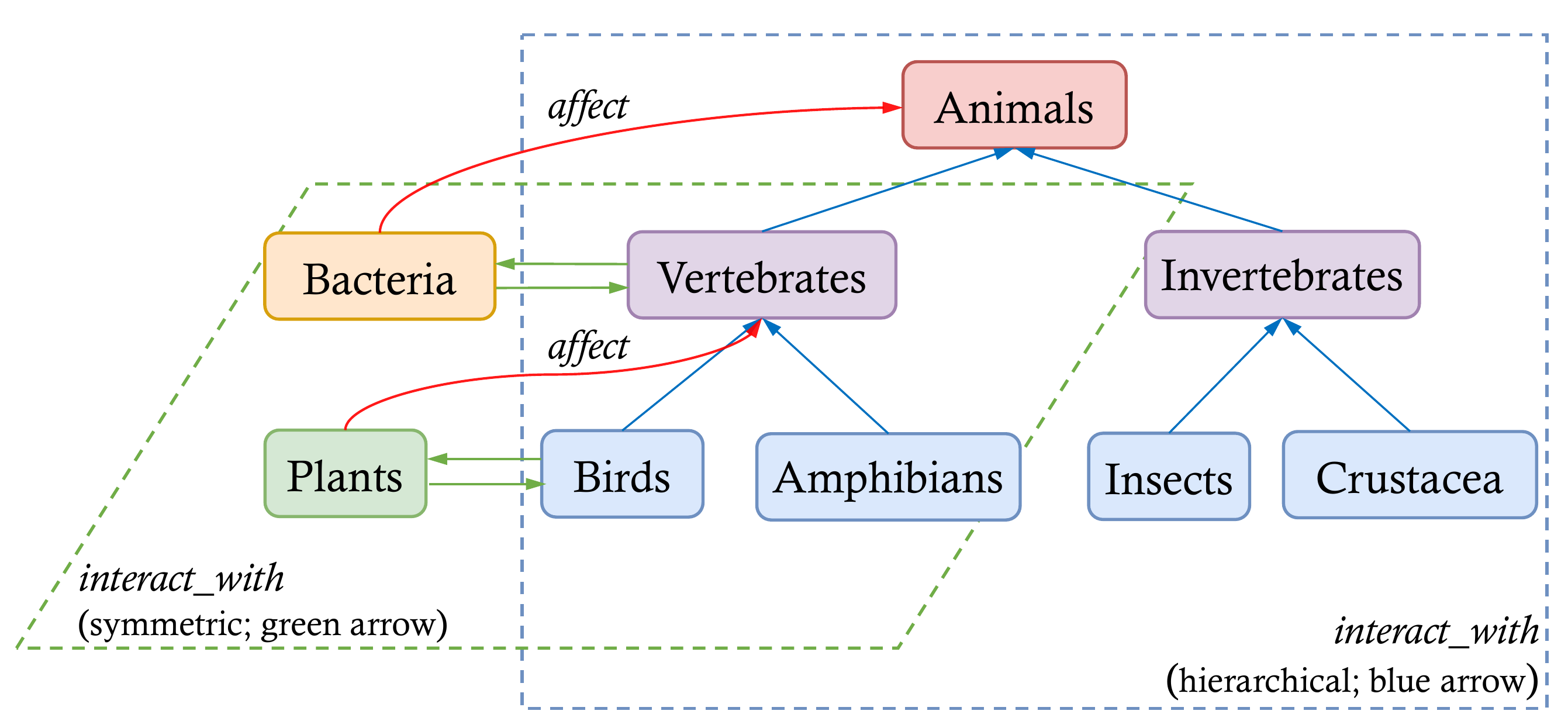} 
\caption{\textit{interacts\_with} is symmetric (green arrows) and hierarchical (blue arrows) in different context. Each \textit{affects} (red arrow) can be composed from a green-arrow relation followed by a blue-arrow one.}
\label{Fig.example} 
\vspace{-\topsep}
\vspace{-5mm}
\end{figure}

Current KGE methods mainly focus on exploiting geometric transformations and embedding spaces 
to model relational patterns such as (anti)symmetry, inversion, and composition. 
TransE~\cite{TransE} represents each relation as a translation from a head entity to a tail entity. With translations alone, TransE cannot model many relation types such as symmetric ones. %
In contrast, RotatE~\cite{RotatE} represents each relation as a rotation in complex space. It proves that it can model (anti)symmetry, inversion, and composition patterns. QuatE~\cite{QuatE} extends RotatE's complex number representation to a hypercomplex number representation.%

A drawback of rotation-based KGE models is that their representations are entrenched in (Euclidean) circular rotation, and hence they are unable to model hierarchical and tree-like structures (e.g. \textit{hypernym} and \textit{part\_of}). Such hierarchical relations are common and even pervasive in some KGs. Since, in circular rotation, all rotating points are constrained to be at the same distance from the center of a circle, it is hard to model relations whose semantics require that entities move at different distances from the nexus. 

To overcome this shortcoming, recent models project KGs into hyperbolic space~\cite{MurP,AttH}. 
The hyperbolic models inevitably lose the basic properties of Euclidean-space transformations, % 
and thus cannot avail themselves of these useful operations.
Moreover, it is difficult to seamlessly integrate the hyperbolic models with extant non-hyperbolic models to create more powerful hybrids because the models' different geometric representations do not cohere.  

What we want is the best of best worlds, i.e., a model capable of effecting both circular rotations and hyperbolic transformations, in a coherent geometric representation. This allows the model to choose the best representation for each relation, e.g., circular rotations for symmetric/inversion relations and hyperbolic rotations for hierarchical patterns. In addition, for relations that exhibit both circular and hyperbolic characteristics (e.g., the \textit{interact\_with} relation %
in Figure~\ref{Fig.example}), the model would rely on the data to choose the sweet spot balancing both transformations. %
For relations that are best captured by the composition of circular and hyperbolic rotations (e.g., \textit{affects} in Figure~\ref{Fig.example}), %
the model would learn the best composition of both representations jointly. Lastly, by subsuming circular rotations, the model would inherit the representational prowess of circular-rotation models.

In this paper, we propose precisely such a model named BiQUE. BiQUE employs a powerful algebraic system called \textit{biquaternions}~\cite{ward1997} to represent KGs. %
Most common number systems used by current KGE methods (including real numbers, complex numbers, and real quaternions) are subsumed and systematically unified by biquaternions. Further, the Hamilton product of biquaternions, at the core of BiQUE, imbues it with a strong geometric interpretation that combines both circular rotations and hyperbolic rotations. In sum, our contributions are as follows. 
\vspace{-\topsep}
\begin{itemize}[leftmargin=3mm]
    \setlength{\itemsep}{0pt}
    \item To our knowledge, we are the first to use biquaternionic algebra for KGEs. 
    Algorithmically, we contribute by designing a flexible score function that leverages multiple geometric transformations (scaling, translation, circular rotation, and hyperbolic rotation).
    \item Theoretically, we contribute by rigorously proving that BiQUE's biquaternionic transformation is equivalent to the composition of a circular rotation and a hyperbolic rotation.
    \item Empirically, we contribute by validating and analyzing BiQUE's effectiveness on five KG benchmarks that span a wide gamut of sizes. %
\end{itemize}

\section{Related Work}
\label{sec:related}

\vspace{-\topsep}
We briefly survey KGE methods that are most relevant to our approach.

\vspace{-\topsep}
\paragraph{Euclidean models.}
These models represent entities and relations by real vectors, and can be categorized into translation-based models~\cite{TransE,TransH,TransR,TransSparse}, semantic-matching models~\cite{RESCAL,DistMult}, and neural models~\cite{ConvE,InteractE}. Euclidean models typically cannot represent all relation types in a KG, e.g., TransE cannot model symmetric patterns, and DistMult cannot model antisymmetric/inversion patterns~\cite{RotatE}. Euclidean models typically need large embedding dimensions for good empirical performance. %

\vspace{-\topsep}
\paragraph{Complex-valued models.}
ComplEx~\cite{ComplEx} uses complex-valued tensor factorization and complex conjugation to model antisymmetric relations. RotatE~\cite{RotatE} models each relation as a 2D circular rotation in a complex vector space. It can model several relation types, but falls short when modeling hierarchical patterns. QuatE~\cite{QuatE} extends RotatE's complex representation to a hypercomplex one via quaternion embeddings, and represents rotations in a four-dimensional real-number space. However, it shares the weakness of circular-rotation methods in being deficient when modeling hierarchical semantics. 
DualE~\cite{DualE} incorporates dual numbers into quaternions, and thereby unifies circular rotation with {\it translation}. This geometric combination of DualE differs from that of our BiQUE model (Section~\ref{sec:bique}), which unifies circular rotation with {\it hyperbolic rotation}. Another difference is that BiQUE models translations via a different mechanism in the form of a relation-specific biquaternion.

\vspace{-\topsep}
\paragraph{Hyperbolic models.}
Recent models have used hyperbolic space because of its amenability to representing hierarchical structures. MurP~\cite{MurP} and ATTH~\cite{AttH} both adopt the Poincar\'e-ball model to represent entities and relations. MurP utilizes Möbius multiplication and addition with relation-specific parameters to transform entities. ATTH integrates hyperbolic rotations and reflections via an attention mechanism, and learns hyperbolic curvatures automatically. Both MurP and ATTH have rigid hyperbolic geometric assumptions, and they remain challenging to optimize directly in hyperbolic space. 

In contrast to existing systems, our model BiQUE overcomes their weaknesses by integrating the strengths of their respective geometric representations into one coherent representation using biquaternions. By subsuming the complex-valued rotation-based models (e.g., complEx and QuatE), it retains their strengths in capturing (anti)symmetric, inversion, and composition patterns. By incorporating a hyperbolic representation, it is also able to model hierarchical semantics.

\section{Background}
\label{sec:background}

Biquaternions, endowed with rich algebraic properties, have been widely used in quantum mechanics, general relativity, and signal processing~\cite{pei2004commutative,GONG2011821}, but have yet to make inroads into knowledge graph embeddings. A biquaternion is defined on a four-dimensional vector space over the field $\bbC$ of complex numbers. We denote a complex number $c \in \bbC$ as $c = c_r + c_i\bfI$ where $c_r, c_i \in \bbR$ are real numbers, and $\bfI$ is the usual imaginary unit ($\bfI^2 = -1$).

\theoremstyle{definition}
\begin{definition}[]
The basic algebraic forms of a \linebreak biquaternion $q$ are 
\vspace{-\topsep}
\begin{flalign}
&\;\;\;\;\;\;\;\;\;\;\;\;\;\;\;\;\;\;\;\; w + x\bfi + y\bfj + z\bfk&& \label{eqn:biquat}\\
&\!=\!(w_r\!+\!w_i\bfI)\!+\!(x_r\!+\!x_i\bfI)\bfi\!+\!(y_r\!+\!y_i\bfI)\bfj\!+\!(z_r\!+\!z_i\bfI)\bfk&& \nonumber \\
&\!=\!q_r\!+\!q_i\bfI&& \label{eqn:qrqi}
\end{flalign}
where $w, x, y, z\!\in\!\bbC$ are $q$'s {\it coefficients}, $w_r, x_r, y_r, z_r, w_i, x_i, y_i, z_i\!\in\!\bbR$, $q_r\!=\!w_r\!+\!x_r\bfi \!+\! y_r\bfj \!+\! z_r\bfk$,  $q_i\!=\!w_i\!+\!x_i\bfi \!+\! y_i\bfj \!+\! z_i\bfk$, and $\bfi, \bfj, \bfk$ are imaginary units that have the following (non)commutative multiplication properties:
\vspace{-\topsep}
\begin{align}
\resizebox{.89\linewidth}{!}{
\begin{math}
\begin{aligned}
\label{eqn:commuteprop}
\bfi^2 = \bfj^2 = \bfk^2 = -1, \;\;\;\;\; &\bfi\bfI=\bfI\bfi, \bfj\bfI =\bfI\bfj,  \bfk\bfI=\bfI\bfk\\
\bfi \bfj = -\bfj \bfi = \bfk, \;\; \bfj \bfk = -\bfk &\bfj =\bfi, \;\; \bfk\bfi=-\bfi\bfk=\bfj, %\\
\end{aligned}
\end{math}
}
\end{align}
\end{definition}

We denote the {\it scalar} and {\it vector} parts of $q$ respectively as $s(q)\!=\!w$ and $v(q)\!=\!x\bfi \!+\! y\bfj \!+\! z\bfk$.
A {\it pure} biquaternion $q$ is one with $s(q)\!=\!0$. 
A {\it quaternion}~\cite{Hamilton1844} is a restricted biquaternion, in which $w, x, y, z\!\in\!\bbR$ 
(e.g., $q_r$ and $q_i$ in Equation~\ref{eqn:qrqi} are quaternions). 
Complex numbers and real numbers are both special cases of biquaternions.

The biquaternion $q\!=\!w \!+\! x\bfi \!+\! y\bfj \!+\! z\bfk$ has several equivalent representations~\cite{ward1997,jafari2016}:
(a) as the vector $\calV(q) = [w, x, y, z]^T$; (b) as $||q|| (\cos \theta \!+\! u \sin \theta)$ (where $\theta \!=\! \cos^{-1}(w/||q||)$, $\theta\! \in\! \bbC$, $u \!=\! v(q)/||v(q)||$, and $||q||\!=\!\sqrt{w^2 \!+\! x^2 \!+\! y^2 \!+\!z^2}$); and (c) as the matrix
\vspace{-\topsep}
\begin{equation}
\label{eqn:matrixrep}
\calM(q) =
\left [
\begin{array}{rrrr}
w \!&\! -x \!&\! -y \!&\! -z\\
x \!&\!  w \!&\!  z \!&\! -y\\
y \!&\! -z \!&\!  w \!&\!  x\\
z \!&\!  y \!&\! -x \!&\!  w\\
\end{array}
\right ].
\end{equation}

Next we present basic operations on a biquaternion $q\!=\!w \!+\! x\bfi \!+\! y\bfj \!+\! z\bfk$. The {\it conjugate} of $q$ is denoted as $\bar{q} = \!w \!-\! x\bfi \!-\! y\bfj \!-\! z\bfk$. The {\it complex conjugate} of $q$ is denoted as $q^* = \!w^* \!+\! x^*\bfi \!+\! y^*\bfj \!+\! z^*\bfk$ ($c^*\!=\!c_r\!-\!c_i \bfI$ is the standard complex conjugate of a complex number $c\!=\!c_r\!+\!c_i \bfI$). 
Two biquaternions  $q_1\!=\!w_1 \!+\! x_1\bfi \!+\! y_1\bfj \!+\! z_1\bfk$ and $q_2\!=\!w_2 \!+\! x_2\bfi \!+\! y_2\bfj \!+\! z_2\bfk$ are added and subtracted (in the obvious manner) as $q_1\!\pm\!q_2\!=\! (w_1\!\pm\! w_2) \!+\! (x_1\!\pm\! x_2)\bfi  \!+\! (y_1\!\pm\! y_2)\bfj \!+\! (z_1\!\pm\! z_2)\bfk$. 
The multiplication $q_1q_2$ between $q_1$ and $q_2$ can be obtained via standard algebraic distributivity (while obeying the properties in Equations~\ref{eqn:commuteprop}, and following the normal multiplication rule of complex numbers for the products between coefficients) as
 \vspace{-\topsep}
\begin{align}
\label{eqn:hamproduct}
q_1q_2 
= & w_1w_2 \!-\! x_1x_2 \!-\! y_1y_2 \!-\! z_1z_2 &&\nonumber \\
   &+(w_1x_2 \!+\! x_1w_2 \!+\! y_1z_2 \!-\! z_1y_2) \bfi &&\nonumber \\
   &+(w_1y_2 \!-\! x_1z_2  \!+\! y_1w_2 \!+\! z_1x_2) \bfj &&\nonumber \\
   &+(w_1z_2 \!+\! x_1y_2 \!-\! y_1x_2 \!+\! z_1w_2) \bfk. &&
\end{align}
Equation~\ref{eqn:hamproduct} is termed the {\it Hamilton product} between $q_1$ amd $q_2$. Alternatively, the multiplication can be equivalently represented as a matrix-vector product
 \vspace{-\topsep}
\begin{align}
\label{eqn:mvproduct}
\calV(q_1q_2) &= \calM(q_2)\,\calV(q_1), %\\
\end{align}
\noindent
or as a matrix-matrix product
\vspace{-\topsep}
\begin{flalign}
\label{eqn:mmproduct}
&\calM(q_1q_2) = \calM(q_2)\,\calM(q_1). %
\end{flalign}
\noindent
Equations~\ref{eqn:mvproduct} and~\ref{eqn:mmproduct} can be easily verified by substituting in Equations~\ref{eqn:matrixrep} and~\ref{eqn:hamproduct}, and using normal matrix multiplication. The set of biquaternions and the set of quaternions are both closed under multiplication, and multiplication is associative but not commutative.
Also note that $\overline{q_1q_2} = \overline{q_2}\;\overline{q_1}$.

By using Equation~\ref{eqn:matrixrep}, we can easily verify that $\calM(\bar{q}) = \calM(q)^T$ and $\calM(q^*) = \calM(q)^*$ (where $\calM(\cdot)^*$ refers to the complex conjugation of each element in the matrix). 

The {\it norm} of $q$ is given by $||q|| \!=\! \sqrt{q\bar{q}} \!=\! \sqrt{\bar{q}q} \!=\!\sqrt{w^2 \!+\! x^2 \!+\! y^2 \!+\!z^2}$ (NB: $||q|| \!=\! ||\bar{q}||$). A {\it unit} biquaternion is one with unit norm, i.e., $||q||=1$.

\section{BiQUE: Biquaternionic Embeddings}
\label{sec:bique}

\subsection{Unification of Circular and Hyperbolic Rotations}
\label{subsec:proof}
We prove that a biquaternion unifies both circular and hyperbolic rotations in $\bbC^4$ space within a single representation in Theorem~\ref{thm:cirhyp} (the proof is in Appendix A). 

\begin{theorem}[]
\label{thm:cirhyp}
Let $\calM(q)$ be the matrix representation of a unit biquaternion $q\!=\!q_r\!+\!q_i \bfI$, where
$q_r\!=\!w_r\!+\!x_r\bfi \!+\! y_r\bfj \!+\! z_r\bfk$,  and $q_i\!=\!w_i\!+\!x_i\bfi \!+\! y_i\bfj \!+\! z_i\bfk$.
$\calM(q)$ can be factorized as $\calM(q) \!=\! \calM(h)\,\calM(u)$ where
$\calM(h)\!=\!$
\vspace{-\topsep}
\begin{flalign*}
&\!\left [\!\!
\begin{array}{rrrr}
\cosh \phi \!&\!- a \bfI\sinh\phi \!&\!- b \bfI\sinh\phi \!&\! - c \bfI\sinh\phi\\
a \bfI\sinh\phi \!&\!  \cosh \phi \!&\!  c \bfI\sinh\phi \!&\! -b \bfI\sinh\phi\\
b \bfI\sinh\phi \!&\! -c \bfI\sinh\phi \!&\!  \cosh \phi \!&\!  a \bfI\sinh\phi\\
c \bfI\sinh\phi \!&\!  b \bfI\sinh\phi \!&\! -a \bfI\sinh\phi \!&\!  \cosh \phi
\end{array}
\!\!\right ]\!\!,
\end{flalign*}
\vspace{-\topsep}
\vspace{-3mm}
\begin{flalign*}
\calM(u)\!=\!\!
\!\left [\!\!
\begin{array}{rrrr}
\cos \theta \!&\!-\frac{x_r \sin \theta}{||v(q_r)||} \!&\!-\frac{y_r \sin \theta}{||v(q_r)||}\!&\! -\frac{z_r \sin \theta}{||v(q_r)||}\\
\frac{x_r \sin \theta}{||v(q_r)||} \!&\!  \cos \theta \!&\!  \frac{z_r \sin \theta}{||v(q_r)||} \!&\! -\frac{y_r \sin \theta}{||v(q_r)||}\\
\frac{y_r \sin \theta}{||v(q_r)||} \!&\! -\frac{z_r \sin \theta}{||v(q_r)||} \!&\!  \cos \theta \!&\!  \frac{x_r \sin \theta}{||v(q_r)||}\\
\frac{z_r \sin \theta}{||v(q_r)||} \!&\!  \frac{y_r \sin \theta}{||v(q_r)||} \!&\! -\frac{x_r \sin \theta}{||v(q_r)||} \!&\!  \cos \theta
\end{array}
\!\!\right ]\!\!,
\end{flalign*}
$\theta\!=\! \cos^{-1} \frac{w_r}{||q_r||}$, $\phi \!=\! \cosh^{-1} ||q_r||$,  $\frac{\overline{q_r}q_i}{||q_r||||q_i||} \!=\! a\bfi \!+\! b\bfj \!+\! c \bfk$, and $\theta, \phi, a,b,c \in \bbR$.
Alternatively, $\calM(q)$ can be factorized as $\calM(q) \!=\! \calM(u)\,\calM(h')$, where $\frac{q_i\overline{q_r}}{||q_i||||q_r||} \!=\! a'\bfi \!+\! b'\bfj \!+\! c' \bfk$, and $\calM(h')\!=\!$
\vspace{-\topsep}
\begin{equation*}
\resizebox{1.05\linewidth}{!}{
\begin{math}
\begin{aligned}
&\left [\!\!
\begin{array}{rrrr}
\cosh \phi \!&\! -a' \bfI\sinh\phi \!&\!- b' \bfI\sinh\phi \!&\! - c' \bfI\sinh\phi\\
a' \bfI\sinh\phi \!&\!  \cosh \phi \!&\!  c' \bfI\sinh\phi \!&\! -b' \bfI\sinh\phi\\
b' \bfI\sinh\phi \!&\! -c' \bfI\sinh\phi \!&\!  \cosh \phi \!&\!  a' \bfI\sinh\phi\\
c' \bfI\sinh\phi \!&\!  b' \bfI\sinh\phi \!&\! -a' \bfI\sinh\phi \!&\!  \cosh \phi
\end{array}
\!\!\right ]\!\!. %
\end{aligned}
\end{math}}
\end{equation*}
In addition, the determinants of {$\calM(h)$, $\calM(h')$ and $\calM(u)$} are 1, and $\calM(h)$, $\calM(h')$ and $\calM(u)$ are orthogonal.
\end{theorem}

From Theorem~\ref{thm:cirhyp}, we know that the matrix $\calM(q)$,
representing a unit biquaternion $q$,
can be expressed as the composition of two matrices $\calM(h)$ and $\calM(u)$
(or $\calM(u)$ and $\calM(h')$). Further, since all elements in $\calM(h)$,
$\calM(h')$, and $\calM(u)$ are derived from $q$, we can construct $\calM(h)$,
$\calM(h')$, and $\calM(u)$ given $q$.

An orthogonal matrix with determinant 1 represents a rotation in the space in which it operates~\cite{artin1957}. Since we know both $\calM(h)$ and $\calM(u)$ are orthogonal and have determinants 1 from Theorem~\ref{thm:cirhyp}, they each represent a rotation in $\bbC^4$ space. From the form of the matrices, we can see that $\calM(u)$ represents a circular rotation, while $\calM(h)$ represents a hyperbolic rotation\footnote{This may be clearer by restricting each matrix to the first two dimensions, which correspond to the square sub-matrix made up of the 4 elements at the top-left corner.}. 
To see the hyperbolic-rotation nature of $\calM(h)$ more clearly, we can use the identities $\cosh \phi \!=\! \cos \bfI\phi$ and $\bfI\sinh \phi \!=\! \sin \bfI\phi$ to represent $\calM(h)$ as
\[
\left [
\begin{array}{rrrr}
\cos \bfI\phi \!&\!- a \sin\bfI\phi \!&\!- b \sin\bfI\phi \!&\! - c \sin\bfI\phi\\
a \sin\bfI\phi \!&\!  \cos \bfI\phi \!&\!  c \sin\bfI\phi \!&\! -b \sin\bfI\phi\\
b \sin\bfI\phi \!&\! -c \sin\bfI\phi \!&\!  \cos \bfI\phi \!&\!  a \sin\bfI\phi\\
c \sin\bfI\phi \!&\!  b \sin\bfI\phi \!&\! -a \sin\bfI\phi \!&\!  \cos \bfI\phi
\end{array}
\right ].
\]
Now $\calM(h)$ takes the form of a ``regular'' rotation matrix (cf. $\calM(u)$), but with a complex angle $\bfI\phi$. According to~\citet{lansey2009visualizing}, a rotation through an imaginary angle $\bfI\phi$ can be understood as a hyperbolic rotation through the real angle $\phi$. Consequently, a unit biquaternion composes these two kinds of rotations in a coherent algebraic representation. (It has been shown by \citet[Corollary 4.1]{jafari2016} that $\calM(q)$ is orthogonal with a determinant of 1, and thus represents an arbitrary rotation in $\bbC^4$. However, that paper does not tease apart the matrix to reveal the contributions of its component circular and hyperbolic rotation matrices like we have done.) 

Our results extend to arbitrary (not necessarily unit) biquaternions. Any biquaternion $q$ is a scaled version of its unit biquaternion, i.e., $q = ||q|| (\frac{q}{||q||})$. Thus its matrix $\calM(q)$ represents a circular rotation followed by a hyperbolic rotation 
(i.e., $\calM(h)\calM(u)$), or a hyperbolic rotation followed by a circular rotation 
(i.e., $\calM(u)\calM(h')$). 
Both rotations are represented by $\frac{q}{||q||}$, followed by a scaling by $||q||$.

We analyze and visualize the $M(u)$ and $M(h)$ rotations in Appendix B.

It is worth noting that the system 
QuatE$^2$~\cite{QuatE}, an experimental baseline in Section~\ref{sec:expts}, 
uses quaternions as its representation. Because quaternions are special cases of biquaternions, QuatE$^2$ only employs the circular rotation matrix $M(u)$ (with its $M(h)$ as the identity matrix). 
Further, note that the power of a biquaternion does not merely comes from doubling the parameters of a quaternion.  A biquaternion achieves better representational power and parameter efficiency by facilitating the interactions between its real and imaginary parameters (see the last paragraph of Section~\ref{sec:B} of the Appendix, and subsection~\ref{subsubsec:complexity}).

\subsection{Problem Definition}
A multi-relational knowledge graph $\calKG$ is represented as a set of directed triples, i.e., $\calKG \!=\! \{(h,r,t)\}$. Each triple $(h,r,t)$ consists of a head entity $h \!\in\! \calE$, a relation $r \!\in\! \calR$, and a tail entity $t \!\in\! \calE$. The numbers of entities and relations are denoted as $|\calE| \!=\! N_e$ and $|\calR| \!=\! N_r$ respectively. The goal of a knowledge graph embedding model is to project entities and relations into a %
continuous vector space while preserving their original semantics. 
The knowledge graph completion (KGC) task requires a model to predict the probability of existence or correctness of unseen triples using  the observed triples $\calKG$.

\subsection{The Proposed Model} 
 
In our BiQUE model, we represent the entities and relations in a $\calKG$ as vectors of biquaternions. Let $\bbQ$ be the set of biquaternions. Each entity $e$ is a vector $Q_e$  of $k$ biquaternions, i.e., $Q_e = [q_1,q_2,\ldots,q_k]^T$, where $q_1,q_2,\ldots,q_k \in \bbQ$. We denote a head entity and a tail entity as $Q_h$ and $Q_t$ respectively. Each relation $r$ is modeled as two vectors $Q_r^+$ and $Q_r^{\times}$, each of which also contains $k$ biquaternions. An entity or relation vector $Q \in \{Q_h, Q_t, Q_r^+, Q_r^{\times}\}$ can also be expressed as  $Q \!=\! w \!+\! x\bfi \!+\! y\bfj \!+\! z\bfk$ where $w, x, y, z \!\in\! \bbC^k$ (i.e., $w, x, y, z$ are each a vector containing $k$ complex numbers, with its $i^{th}$ element corresponding to the $i^{th}$ biquaternion in $Q$).
(Note the similarity between the form of $Q$ and that of a biquaternion in Equation~\ref{eqn:biquat}.) Because a complex number can be represented by two real numbers (its real and imaginary components), $Q$ can be represented with $k\times4\times2 =8k$ real numbers ($8k$ is its {\it embedding size}).
(For expository convenience, we refer to $Q$ as a ``biquaternion'' or an ``embedding''; their structures should be clear from their contexts.)

Currently, the loss functions of KGE models can be roughly categorized as additive and multiplicative ones depending on the relation transformation projecting a head entity to a tail entity. \citet{EXPKG} has recently shown that projections require matrix multiplication, and cannot be achieved via addition alone. Thus, it is necessary to combine both additive and multiplicative operations into a loss function to represent powerful projections.

We represent the transformation due to relation $r$ with the biquaternions $Q_r^{+}$ and $Q_r^{\times}$. 
The embedding $Q_r^{+}$ applies a relation-specific {\em translation} to a head entity's embedding $Q_h$. We realize it by the element-wise addition of biquaternions
(similar to what we do with real vectors for translation):  
\begin{align}
\label{eq:trans}
\begin{aligned}
    Q_{h,r}' &= Q_h + Q_r^{+} \\
    &= (w_h+w_r^{+}) + (x_h+x_r^{+})\bfi \\
    & \ \ \ \ \ \ \ + (y_h+y_r^{+})\bfj + (z_h+z_r^{+})\bfk\\
    &= w'+x'\bfi+y'\bfj+z'\bfk
\end{aligned}
\end{align}
Next the embedding $Q_r^{\times}$ applies a relation-specific multiplicative transformation to the translated head entity $Q_{h,r}'$. The multiplicative transformation is defined via the Hamilton product of biquaternions (Equation~\ref{eqn:hamproduct}) as follows.
\begin{equation}
\label{eq:model}
\resizebox{.86\linewidth}{!}{$\begin{aligned}
 & \widehat{Q_{h,r}} =  Q_{h,r}' \circledast Q_r^{\times} =\\
&(w'\otimes w_r^{\times} - x'\otimes x_r^{\times} - y'\otimes y_r^{\times} - z'\otimes z_r^{\times})+ \\
& (w'\otimes x_r^{\times} + x'\otimes w_r^{\times} + y'\otimes z_r^{\times} - z'\otimes y_r^{\times})\mathbf{i} +\\
&(w'\otimes y_r^{\times} - x'\otimes z_r^{\times} + y'\otimes w_r^{\times} + z'\otimes x_r^{\times})\mathbf{j}+\\
&(w'\otimes z_r^{\times} + x'\otimes y_r^{\times} - y'\otimes x_r^{\times} + z'\otimes w_r^{\times})\mathbf{k}
\end{aligned}$}
\end{equation}
where $\circledast$ denotes the element-wise application of the Hamilton product between $Q_{h,r}'$ and $Q_r^{\times}$, and $\otimes$ denotes the element-wise multiplication between vectors of complex numbers. 
As shown in subsection~\ref{subsec:proof}, each biquaternion in $Q_r^{\times}$ represents a composition of circular rotation,  hyperbolic rotation, and scaling. In the above Hamilton product, we bring this powerful composition to bear on the projection of the translated head entity.
The Hamilton product of biquaternions in Equation~\ref{eq:model} has an added benefit of increasing the potential interaction between entities and relations through the multiplications between different components of the entities and relations (observe that each element in $\{w', x', y', z'\}$ is multiplied with each element in $\{w_r^{\times}, x_r^{\times}, y_r^{\times}, z_r^{\times}\}$.)

Overall, our model %
unifies multiple expressive geometric transformations (translation, scaling,  circular rotation, and hyperbolic rotation) into one coherent representation system. 

\subsection{Score Function and Training Loss}
We measure the plausibility score of a given triple $(h,r,t)$ by computing the vector similarity between the transformed head entity $\widehat{Q_{h,r}}\!=\!\widehat{w}\!+\!\widehat{x}\bfi\!+\!\widehat{y}\bfj\!+\!\widehat{z}\bfk$ (from Equation~\ref{eq:model}), and a candidate tail entity $Q_t\!=\! w_t \!+\! x_t\bfi \!+\! y_t\bfj \!+\! z_t\bfk$ as \begin{align}
\begin{aligned}
    & \ \ \ \ \ f(h,r,t) =\widehat{Q_{h,r}} \cdot Q_t \\%
    &= \langle \widehat{w}, w_t\rangle + \langle \widehat{x}, x_t\rangle + \langle \widehat{y}, y_t\rangle + \langle \widehat{z}, z_t\rangle, 
\end{aligned}
\end{align}
where $\langle\cdot,\cdot\rangle$ denotes the standard dot-product between vectors.

We regard the task of knowledge graph completion as a multi-class classification problem and employ the cross-entropy loss to train our model.
\begin{align}
\mathcal{L} = \sum_{(h,r,t)\in\mathcal{KG}} &\sum_{t' \in \mathcal{E}} \log \big (1 \!+\! \exp\big (y_t f(h,r,t') \big )\big ) \!+\! \Omega \notag\\
y_t &= 
\left\{
\begin{array}{rl}
-1, & t' = t \\
1,  & \text{otherwise}.
\end{array}
\right.
\end{align}

To combat overfitting, we follow previous work~\cite{QuatE}, and append a N3 regularization norm~\cite{N3} to our loss function, thus obtaining
\begin{equation}
\label{eqn:objfunc}
\resizebox{.86\linewidth}{!}{
\begin{math}
\begin{aligned}
    \Omega = \!\!\!\! \sum_{\{h,r,t\}\in \mathcal{KG}} \! \!\!\lambda \big (\lambda_1(\|{Q_h}\|_3^3 \!+\! \|{Q_t}\|_3^3) \!+\! \lambda_2 \|{Q_r}\|_3^3 \big )  
\end{aligned}
\end{math}}
\end{equation}
where $\lambda, \lambda_1, \lambda_2$ are the global, entity and relation regularization hyperparameters respectively. $\|\cdot\|_3$ denotes $L_3$ norm of vectors.

\section{Experiments}
\label{sec:expts}

To validate BiQUE's effectiveness, we conduct extensive experiments on the knowledge graph completion (KGC) task. We use three standard knowledge graph datasets, viz., WN18RR, FB15K-237, and YAGO3-10. In addition, to demonstrate BiQUE's scalability, we run it on two huge commonsense knowledge graph datasets, viz., Concept100k and ATOMIC. Our codes and datasets are publicly available at \url{https://github.com/guojiapub/BiQUE}.

\subsection{Datasets}
The WN18RR~\cite{ConvE} and FB15K-237~\cite{FB15K237} datasets are respectively subsets of WN18 and FB15K (both from~\citet{TransE}).
(Both WN18 and FB15K have test leakage problems, which allow their test triples to be easily inferred. Thus, KGE models typically perform well on those two datasets, and they do not help to differentiate between models. Because of this, we do not use them in our experiments.)
To make the KGC task more challenging, FB15K-237 and WN18RR remove the inverse relations from the original validation and test sets of WN18 and FB15K. 
The CN-100K~\cite{Concept100k} and ATOMIC~\cite{ATOMIC} datasets are two large  knowledge graph benchmarks recently adopted for evaluating commonsense reasoning. ATOMIC mainly describes the reactions, effects, and intents of human behaviors, 
and represents each entity as a phrase with an average length of 4.4 words. CN-100K contains general commonsense knowledge about the world. For CN-100K and ATOMIC, we use the data splits of previous work~\cite{Malaviya}. Table~\ref{tab:datastats} provides details on the datasets. (Note that the datasets span a wide range of sizes.)

Both WN18RR and YAGO3-10 contain many relations with hierarchical semantics, e.g., \textit{hypernym} and \emph{part\_of}~\cite{AttH}. On the other hand, most of FB15K-237's edges are antisymmetric, and it does not have much hierarchical structure~\cite{MurP}. The varying level of hierarchical structure in the datasets helps to highlight  BiQUE's adaptability to datasets with different relation types.
ATOMIC mainly contains cause-effect relations that are not hierarchical. CN100k contains several hierarchical relations (e.g., \textit{IsA} and \textit{AtLocation}). Aside from their large sizes, these two datasets have the challenging feature of being extremely sparse.

\begin{table}[htbp]
\centering
\scalebox{0.70}{
\begin{tabular}{lrrrrr}
\toprule
Dataset  & \#Entities & \#Relations & \#Train & \#Valid & \#Test \\
\midrule
FB15K-237 & 14,541  & 237   & 272,115   & 17,535    & 20,466  \\
WN18RR    & 40,943  & 11    & 86,835    & 3,034     & 3,134   \\
YAGO3-10  & 123,188 & 37    & 1,079,040 & 5,000     & 5,000   \\
\midrule
CN-100K   & 78334   & 34  & 100,000   & 1,200    & 1,200  \\
ATOMIC    & 304,388 & 9   & 610,536   & 87,700   & 87,701  \\
\bottomrule
\end{tabular}}
\caption{Knowledge Graph Benchmarks.}
\label{tab:datastats}
\end{table}
\subsection{Baselines}
We compare our model to strong baselines that operate in different geometric spaces. For Euclidean space, we use TransE~\cite{TransE}, DistMult~\cite{DistMult}, ConvE~\cite{ConvE}, InteractE~\cite{InteractE}, and CompGCN~\cite{CompGCN}. For complex-valued space, we use ComplEx~\cite{ComplEx}, RotatE~\cite{RotatE}, QuatE$^2$~\cite{QuatE} (the version with N3 regularization and reciprocal learning) and DualE$^1$(without type constraints)~\cite{DualE}. For hyperbolic space, we use MurP~\cite{MurP} and ATTH~\cite{AttH}. For the commonsense datasets, we also include ConvTransE~\cite{ConvTransE}.

\subsection{Evaluation Protocol}
We use standard evaluation metrics for the knowledge graph completion (KGC) task, viz., mean reciprocal rank (MRR) and Hits$@k$ with cut-off values $k \in \{1,3,10\}$. For both MRR and Hits$@k$, the larger the metric, the better the performance of a model. We adopt the \texttt{BOTTOM} setting~\cite{re-evaluation} when ranking candidate triples and we consistently apply it to our BiQUE model, i.e., the correct triple is always inserted at the end of a list of triples with the same plausibility scores. This is the strictest evaluation protocol for KGC tasks, and provides the best reflection of a model's performance. Finally, we report filtered results like previous work~\cite{TransE} for fair comparisons. (Implementation details are in the appendix.)

\begin{table*}[ht!]
\centering
\scalebox{0.87}{
\begin{tabular}{lrrrrrrrrrrrr}
\toprule
 &  \multicolumn{4}{c}{WN18RR}  &  \multicolumn{4}{c}{FB15K-237} & \multicolumn{4}{c}{YAGO3-10} \\
\cmidrule(lr){2-5} \cmidrule(lr){6-9} \cmidrule(lr){10-13}
Models  & MRR & H@1 & H@3 & H@10 & MRR & H@1 & H@3 & H@10 & MRR & H@1 & H@3 & H@10 \\
\midrule
TransE & 0.226 & - & - & 0.501 & 0.294 & - & - & 0.465 & - & - & - & - \\
DistMult & 0.430 & 0.390 & 0.440 & 0.490 & 0.241 & 0.155 & 0.263 & 0.419 & 0.340 & 0.240 & 0.380 & 0.540 \\
ConvE & 0.430 & 0.400 & 0.440 & 0.520 & 0.325 & 0.237 & 0.356 & 0.501 & 0.440 & 0.350 & 0.490 & 0.620 \\
InteractE & 0.463 & 0.430 & - & 0.528 & 0.354 & 0.263 & - & 0.535 & 0.541 & 0.462 & - & 0.687 \\
CompGCN & 0.479 & \underline{0.443} & 0.494 & 0.546 & 0.355 & 0.264 & 0.390 & 0.535 & - & - & - & -\\
\midrule
ComplEx-N3 & 0.480 & 0.435 & 0.495 & 0.572 & 0.357 & 0.264 & \underline{0.392} & 0.547 & \underline{0.569} & \underline{0.498} & 0.609 & 0.701 \\
RotatE & 0.476 & 0.428 & 0.492 & 0.571 & 0.338 & 0.241 & 0.375 & 0.533 & 0.495 & 0.402 & 0.550 & 0.670 \\
QuatE$^2$ & 0.482 & 0.436 & \underline{0.499} & 0.572 & \textbf{0.366} & \textbf{0.271} & \textbf{0.401} & \textbf{0.556} & 0.568 & 0.493 & 0.611 & \underline{0.706}\\
DualE$^1$ & 0.482 & 0.440 & 0.500 & 0.561 & 0.330 & 0.237 & 0.363 & 0.518 & - & - & - & - \\
\midrule
MurP & 0.481 & 0.440 & 0.495 & 0.566 & 0.335 & 0.243 & 0.367 & 0.518 & 0.354 & 0.249 & 0.400 & 0.567 \\
ATTH & \underline{0.486} & \underline{0.443} & \underline{0.499} & \underline{0.573} & 0.348 & 0.252 & 0.384 & 0.540 & 0.568 & 0.493 & \underline{0.612} & 0.702 \\
\midrule
BiQUE & \textbf{0.504} & \textbf{0.459} & \textbf{0.519} & \textbf{0.588} & \underline{0.365} & \underline{0.270} & \textbf{0.401} & \underline{0.555} & \textbf{0.581} & \textbf{0.509} & \textbf{0.624} & \textbf{0.713}\\
\bottomrule
\end{tabular}}
\caption{Best results are \textbf{bolded}, and second best results are \underline{underlined}. Results for DistMult, ConvE, ComplEx-N3, RotatE, MurP, and ATTH are from~\citet{AttH}. Results for TransE and QuatE$^2$ are from~\citet{QuatE}. Results for InteractE, CompGCN and DualE$^1$ are from their original papers. YAGO3 results for QuatE$^2$ are obtained using our implementation (QuatE$^2$ can be viewed as a special case of BiQUE 
as discussed in Section~\ref{subsec:proof}).} 
\label{tab:main}
\end{table*}

\begin{table*}[ht!]
\centering
\scalebox{0.87}{
\begin{tabular}{lrrrrrrrr}
\toprule
& \multicolumn{4}{c}{CN-100K} & \multicolumn{4}{c}{ATOMIC} \\
\cmidrule(lr){2-5} \cmidrule(lr){6-9} 
Models  & MRR  & H@1 & H@3 & H@10 & MRR & H@1 & H@3 & H@10   \\
\midrule
DistMult         & 0.090 & 0.045  & 0.098 & 0.174 & 0.124 & 0.092  & 0.152 & 0.183 \\
ComplEx          & 0.114 & 0.074  & 0.125 & 0.190 & 0.142 & 0.133 & 0.141 & 0.160 \\
ConvE            & 0.209 & 0.140  & 0.229 & 0.340 & 0.101 & 0.082  & 0.103 & 0.134 \\
RotatE          & 0.247 & -      & 0.282 & 0.454 & 0.112 & -     & 0.115 & 0.156 \\
ConvTransE  & 0.187  & 0.079 & 0.239 & 0.390 & 0.129 &  0.129 & 0.130 &  0.130 \\
QuatE$^2$ & \underline{0.313} & \textbf{0.217} & \underline{0.356} & \underline{0.504} & \underline{0.187} & \underline{0.167} & \underline{0.191} & \underline{0.225}  \\
\midrule
BiQUE     & \textbf{0.320} & \underline{0.216}  &  \textbf{0.359} &  \textbf{0.553} & \textbf{0.191} & \textbf{0.171} & \textbf{0.196} & \textbf{0.230}\\
\bottomrule
\end{tabular}}
\caption{Best results are \textbf{bolded}, and second best results are \underline{underlined}. Results for DistMult, ComplEx, ConvE, and ConvTransE are from~\citet{Malaviya}. Results for RotatE are from~\citet{InductivE}. Results for QuatE$^2$ are obtained using our implementation (QuatE$^2$ is a special case of BiQUE
as discussed in Section~\ref{subsec:proof}).}
\label{tab:main2}
\end{table*}

\subsection{Results}
Table~\ref{tab:main} shows the experimental results. Following standard practice adopted by our comparison systems, we show the best results for BiQUE. (We pick the best result over 10 runs with different random initializations. The appendix contains the average scores over the runs, and the standard deviations.)

From Table~\ref{tab:main}, we see that BiQUE is the best performer on four of the five datasets, and is a close second on the remaining dataset.
On the two hierarchical datasets WN18RR and YAGO3-10, BiQUE achieves new state-of-the-art results on all metrics, and surpasses the second best models by a clear margin. WN18RR contains a large proportion of symmetric relations (which are amenable to being modeled with circular rotations) and hierarchical relations (which are amenable to being modeled with hyperbolic rotations). BiQUE's good performance on this dataset provides evidence that BiQUE's composition of circular and hyperbolic rotations is useful in modeling these disparate relation types simultaneously.  
BiQUE also consistently outperforms the hyperbolic models MurP and AttH on all metrics. 

On FB15K-237, BiQUE is second best; however, BiQUE's scores are only marginally lower than those of the best system QuatE$^2$. %
As observed by \citet{MurP}, the vast majority of relations in FB15K-237 do not form hierarchies. Consequently, BiQUE's hyperbolic transformation does not play a principal role on FB15K-237, %
and BiQUE falls back on only using its circular rotation transformation. %
QuatE$^2$ can be viewed as a special case of BiQUE that only has circular rotations ($\calM(u)$ in Thoerem~\ref{thm:cirhyp}). Since both use the same transformation, their results are (almost) indistinguishable on FB15K-237.

Table~\ref{tab:main2} shows the results on the large commonsense graphs, CN-100K and ATOMIC. We see that these datasets are a lot more challenging with many KGE models having MRRs below or hovering around 0.3. (Because of the datasets' large sizes, many extant KGE models do not experiment on them, and we compare against the models that have been reported in the literature. \citet{Malaviya} describes systems that use BERT embeddings, and thus encapsulate a lot of commonsense prior knowledge; for a fair comparison, we do not use such systems as baselines.)
Table~\ref{tab:main2} shows that BiQUE outperforms the previously reported state-of-the-art results on CN-100K (RotatE) and ATOMIC (ComplEx) (by 29.6\% and 34.5\% on MRR respectively). Further, by composing hyperbolic rotation with circular rotation, BiQUE surpasses QuatE$^2$ that only uses the latter rotation.

\begin{table}[t]
\centering
\scalebox{0.68}{
\setlength{\tabcolsep}{1.3mm}{
\begin{tabular}{lrrrrr}
\toprule
Relation Name                   &  \#Tripes & RotH & QuatE$^2$    & BiQUE    & Lift\\
\midrule
hypernym                        &  1,251  & 0.276 & \underline{0.283}  & \textbf{0.306}  & 8.13\% \\
derivationally\_related\_form   &  1,074  & 0.968 & \underline{0.969}  & \textbf{0.970}  & 0.10\%\\
instance\_hypernym              &  122    & 0.520 & \underline{0.549}  & \textbf{0.602}  & 9.65\%\\
also\_see                       &  56     & \underline{0.705} & 0.688  & \textbf{0.750}  & 6.38\%\\
member\_meronym                 &  253    & \underline{0.399} & 0.389  & \textbf{0.453}  & 13.53\%\\
synset\_domain\_topic\_of       &  114    & 0.447 & \underline{0.518}  & \textbf{0.539}  & 4.05\%\\
has\_part                       &  172    & \underline{0.346} & 0.337  & \textbf{0.392}  & 13.29\%\\
member\_of\_domain\_usage       &  24     & \underline{0.438} & \textbf{0.563}  & \textbf{0.563}  & -\\
member\_of\_domain\_region      &  26     & \underline{0.365} & 0.327  & \textbf{0.500}  & 36.99\%\\
verb\_group                     &  39     & \textbf{0.974} &  \textbf{0.974} & \textbf{0.974}   & - \\
similar\_to                     &  3      & \textbf{1.000} & \textbf{1.000}  & \textbf{1.000}   & -\\
\bottomrule
\end{tabular}}}
\caption{Best results are \textbf{bolded}, and second best results are \underline{underlined}. Comparison of H@10 per relation for BiQUE and baselines on WN18RR. Results for RotH are from~\citet{AttH}. QuatE$^2$'s results are obtained by running its official code
(\url{https://github.com/cheungdaven/QuatE})}
\label{tab:relanalysis}
\end{table}

\subsection{Analysis}

\subsubsection{Performance per Relation}

To provide a fine-grained analysis of BiQUE's results, we report its performance per relation on WN18RR in Table~\ref{tab:relanalysis}. 
A large portion of the WN18RR dataset consists of hierarchical triples, such as \textit{hypernym} and \textit{instance\_hypernym}, which account for more than 43\% of training examples. We see that BiQUE achieves the best performance on all 11 relation types compared with current top-performing models. BiQUE not only performs well on hierarchical and tree-like relations (e.g., \textit{hypernym} and \textit{instance\_hypernym}), but also obtains significant improvements on challenging one-to-many relations  %
(e.g., \textit{member\_of\_domain\_region} and \textit{member\_meronym}). It is worth noting that BiQUE does not sacrifice its performance on other relation types for the abovementioned improvements. In fact, BiQUE also achieves the best performance for symmetric relations (e.g., \textit{derivationally\_related\_form} and \textit{verb\_group}). All in all, these fine-grained results support our hypothesis that BiQUE's integration of multiple geometric transformations allows it to make good trade-offs among various representations for different relation types, thereby allowing it to pick the best one (or combinations thereof) for optimal performance.

\subsubsection{Model Variants and Ablation Study}
\label{sec:variants}
Table~\ref{tab:variants} shows the performances of variants and ablations of BiQUE. %
We test the impact of BiQUE's scaling operation by normalizing the relation rotation biquaternion $Q_r^{\times}$ in two ways: $Q_r^{\times\star}$ uses the regular normalization of real vectors, and $Q_r^{\times\triangleleft}$ uses the standard normalization of biquaternions. (See appendix for details.) 
Compared to BiQUE, both normalized variants perform worse. Thus the norm of
$Q_r^{\times}$ plays an important role in having a scaling effect. Further, from our ablation study, we observe that without the translation operator $Q_r^+$, BiQUE performs worse on both datasets. % 
This shows that rotations cannot fully replace translations in KGE models. Table~\ref{tab:variants} also shows that regularization is important for BiQUE to avoid overfitting.  

\begin{table}[t]
\centering
\scalebox{0.8}{
\begin{tabular}{lrrrr}
\toprule
\multirow{2}*{Variants}   & \multicolumn{2}{c}{WN18RR} & \multicolumn{2}{c}{FB15K-237}  \\
\cmidrule(lr){2-3} \cmidrule(lr){4-5} 
  & MRR & H@3 & MRR & H@3\\
\midrule
BiQUE & 0.504 & 0.519 & 0.365 & 0.401\\
$\big ((Q_h + Q_r^{+}) \circledast Q_r^{\times\star}\big ) \cdot Q_t$ & 0.491 & 0.502 & 0.351 & 0.384 \\
$\big ((Q_h + Q_r^{+}) \circledast Q_r^{\times\triangleleft}\big ) \cdot Q_t$ & 0.486 & 0.500 & 0.350 & 0.385 \\
\midrule
w/o $Q_r^{+}$ & 0.490 & 0.505 & 0.362 & 0.398  \\
w/o regularizer & 0.457 & 0.467 & 0.344 & 0.378\\
\bottomrule
\end{tabular}}
\caption{Variants and ablations of BiQUE on WN18RR.}
\label{tab:variants}
\end{table}

\subsubsection{Model Efficiency}
\label{subsubsec:complexity}

\begin{figure}[t]
\centering
\includegraphics[width=0.4\textwidth]{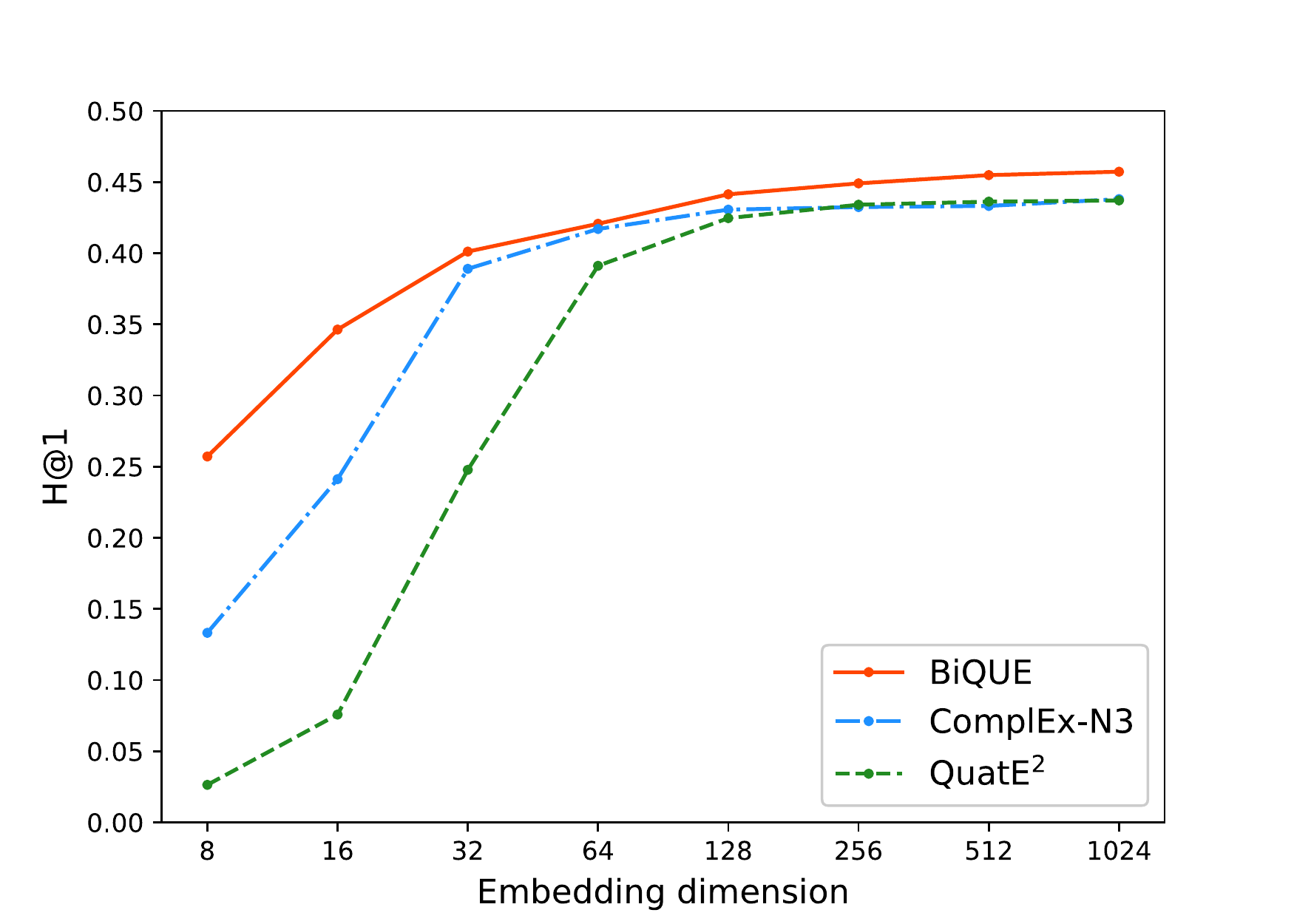} 
\caption{Effect of varying embedding size on performance.}
\label{Fig.dim} 
\end{figure}

We investigate the impact of varying embedding size on performance (H$@$1).
In Figure~\ref{Fig.dim}, the parameters of all systems are tuned, and the results are averaged over 5 runs (with different random initializations). We see that BiQUE's results consistently surpasses those of the strong baselines across embedding dimensions. The disparity is most apparent in the regime of small embedding sizes. This suggests that BiQUE's representation is more effective at modeling the data (and thus do better even with smaller embeddings). 

In Table~\ref{tab:param}, we use the same number of parameters for BiQUE and QuatE$^2$ (the baseline most similar to BiQUE), and show that BiQUE performs better (higher MRR and H$@$3) with fewer epochs. This again supports our hypothesis that BiQUE models the data better (with the same number of parameters) than QuatE$^2$, and thus requires fewer epochs to achieve better results.

\begin{table}[t]
	\centering
    \begin{adjustbox}{max width=0.48\textwidth}
    \setlength{\tabcolsep}{0.9mm}{
	\begin{tabular}{ccccccccc}
		\toprule
		 & \multicolumn{4}{c}{WN18RR} & \multicolumn{4}{c}{FB15K-237} \\
		 \cmidrule(lr){2-5} \cmidrule(lr){6-9} 
		 & \#Params & \#Epoch & MRR & H@3  &  \#params  & \#Epoch & MRR & H@3 \\ \midrule
		QuatE$^2$ & \multirow{2}*{20.97M} & 40000 & 0.481 & 0.496 & \multirow{2}*{7.57M} &  15000 & 0.331 & 0.363\\ 
		BiQUE &  & \textbf{200} & \textbf{0.499} & \textbf{0.515} &  & \textbf{300} & \textbf{0.359} & \textbf{0.397}\\ \bottomrule
	\end{tabular}}
	\end{adjustbox}
	\caption{Performance on WN18RR and FB15K-237 with the same parameter size.}
	\label{tab:param}
\end{table}

\section{Conclusion}
\label{sec:conclu}

In this paper, we propose BiQUE, a novel model that uses biquaternionic algebra for KGEs, and combines multiple geometric transformations in a coherent representation. Our experimental results and detailed empirical analysis demonstrate the effectiveness, scalability, and advantages of our model.  As future work, we will extend BiQUE to work on knowledge {\it hyper}graphs.
\section*{Acknowledgements} 
\vspace{-3mm}
This research is partly supported by MOE's AcRF Tier 1 Grant %
to Stanley Kok. Any opinions, findings, conclusions, or recommendations expressed herein
are solely those of the authors.%

\bibliography{custom}
\bibliographystyle{acl_natbib}

\clearpage
\appendix
\section{Proofs}
We prove that a biquaternion unifies both circular and hyperbolic rotations in $\bbC^4$ space within a single representation in Theorem~\ref{thm:cirhyp}. To do so, we require Theorem~\ref{thm:jafari_det} that is proved by~\citet{jafari2016}, and the definitions covered in Section~\ref{sec:background}. We also prove auxiliary Lemma~\ref{lem:pure}.

\begin{customthm}{A.1}[Jafari, 2016, Theorem 4.1(vi)]
\label{thm:jafari_det}
If $\calM(q)$ is the matrix representation of a biquaternion $q$, then the matrix's determinant is given by $\det[\calM(q)] = ||q||^4$. 
\end{customthm}

\begin{customlem}{A.2}[]
\label{lem:pure}
If $q \!=\! q_r \!+\! q_i \bfI$ is a unit biquaternion (i.e., $||q||\!=\!1$) where $q_r\!=\!w_r\!+\!x_r\bfi \!+\! y_r\bfj \!+\! z_r\bfk$,  and 
$q_i\!=\!w_i\!+\!x_i\bfi \!+\! y_i\bfj \!+\! z_i\bfk$, then $\overline{q_r} q_i$ and $q_i\overline{q_r}$ are pure quaternions (i.e., their scalar parts $s(\overline{q_r} q_i) \!=\! s(q_i\overline{q_r}) \!=\! 0$), $||q_r|| \!=\! \cosh \phi$ and $||q_i|| \!=\! \sinh \phi\;$, where $\phi \in \bbR$.
\end{customlem}
\vspace{-\topsep}
\begin{proof} 
Note that both $q_r$ and $q_i$ are quaternions, and $q\!=\!(w_r\!+\!w_i\bfI)\!+\!(x_r\!+\!x_i\bfI)\bfi\!+\!(y_r\!+\!y_i\bfI)\bfj\!+\!(z_r\!+\!z_i\bfI)\bfk$.
\vspace{-\topsep}
\begin{flalign*}
||q||^2\!=& (w_r\!+\!w_i\bfI)^2\!+\!(x_r\!+\!x_i\bfI)^2\!+\!(y_r\!+\!y_i\bfI)^2 &&\\
             &+\!(z_r\!+\!z_i\bfI)^2 &&\\
          \!=&(w_r^2\!-\!w_i^2 + x_r^2\!-\!x_i^2 + y_r^2\!-\!y_i^2 + z_r^2\!-\!z_i^2) &&\\
          &+ 2 (w_rw_i+x_rx_i+y_ry_i+z_rz_i)\bfI. &&
\end{flalign*}
Since $||q||^2 = 1$ is real, we know that the above imaginary part  $w_rw_i+x_rx_i+y_ry_i+z_rz_i = 0$.
\vspace{-\topsep}
\begin{flalign*}
s(\overline{q_r}q_i)\!= & s((w_r\!-\!x_r\bfi\!-\!y_r\bfj\!-\!z_r\bfk) \cdot&&\\
&\;\;\;\;\;\;\;(w_i\!+\!x_i\bfi\!+\!y_i\bfj\!+\!z_i\bfk))\\
=&\; w_rw_i+x_rx_i+y_ry_i+z_rz_i \;\; \mbox{(using Eq.~\ref{eqn:hamproduct})} \\
=&\; 0. \\%
s(q_i\overline{q_r})\!=& s((w_i\!+\!x_i\bfi\!+\!y_i\bfj\!+\!z_i\bfk) \cdot &&\\
&\;\;\;\;\;\;\;(w_r\!-\!x_r\bfi\!-\!y_r\bfj\!-\!z_r\bfk))\\
=&\; w_iw_r+x_ix_r+y_iy_r+z_iz_r \;\; \mbox{(using Eq.~\ref{eqn:hamproduct})} \\
=&\; 0.
\end{flalign*}
Thus $\overline{q_r} q_i$ and $q_i\overline{q_r}$ are both pure quaternions (recall that the set of quaternions is closed under multiplication).
\vspace{-\topsep}
\begin{flalign*}
&||q||^2\!=\! \bar{q}q \!=\! \overline{(q_r \!+\! q_i\bfI)}(q_r \!+\! q_i\bfI) \!=\! (\overline{q_r} \!+\!\overline{q_i}\bfI)(q_r \!+\! q_i\bfI) && \\
&\!=\!\overline{q_r}q_r + \overline{q_i}q_i\bfI^2 + (\overline{q_r}q_i + \overline{q_i}q_r) \bfI \\
&\!=\! ||q_r||^2 - ||q_i||^2 + (\overline{q_r}q_i + \overline{q_i}q_r) \bfI.
\end{flalign*}
Since $||q||^2\!=\!1$ is real, 
$\overline{q_r}q_i \!+\! \overline{q_i}q_r\!=\!0$ and
$||q_r||^2 \!-\! ||q_i||^2\!=\!1$. The latter is the equation of a hyperbola. Since $||q_r||$ is real, there must exist $\phi \in \bbR$ such that  $||q_r|| = \cosh \phi$ and $||q_i|| = \sinh \phi$ (which gives the identity $\cosh^2 \phi - \sinh^2 \phi = 1$).
\end{proof}

\begin{customthm}{4.1}
Let $\calM(q)$ be the matrix representation of a unit biquaternion $q\!=\!q_r\!+\!q_i \bfI$, where
$q_r\!=\!w_r\!+\!x_r\bfi \!+\! y_r\bfj \!+\! z_r\bfk$,  and $q_i\!=\!w_i\!+\!x_i\bfi \!+\! y_i\bfj \!+\! z_i\bfk$.
$\calM(q)$ can be factorized as $\calM(q) \!=\! \calM(h)\,\calM(u)$ where
$\calM(h)\!=\!$
\vspace{-\topsep}
\begin{flalign*}
&\!\left [\!\!
\begin{array}{rrrr}
\cosh \phi \!&\!- a \bfI\sinh\phi \!&\!- b \bfI\sinh\phi \!&\! - c \bfI\sinh\phi\\
a \bfI\sinh\phi \!&\!  \cosh \phi \!&\!  c \bfI\sinh\phi \!&\! -b \bfI\sinh\phi\\
b \bfI\sinh\phi \!&\! -c \bfI\sinh\phi \!&\!  \cosh \phi \!&\!  a \bfI\sinh\phi\\
c \bfI\sinh\phi \!&\!  b \bfI\sinh\phi \!&\! -a \bfI\sinh\phi \!&\!  \cosh \phi
\end{array}
\!\!\right ]\!\!,
\end{flalign*}
\vspace{-\topsep}
\vspace{-3mm}
\begin{flalign*}
\calM(u)\!=\!\!
\!\left [\!\!
\begin{array}{rrrr}
\cos \theta \!&\!-\frac{x_r \sin \theta}{||v(q_r)||} \!&\!-\frac{y_r \sin \theta}{||v(q_r)||}\!&\! -\frac{z_r \sin \theta}{||v(q_r)||}\\
\frac{x_r \sin \theta}{||v(q_r)||} \!&\!  \cos \theta \!&\!  \frac{z_r \sin \theta}{||v(q_r)||} \!&\! -\frac{y_r \sin \theta}{||v(q_r)||}\\
\frac{y_r \sin \theta}{||v(q_r)||} \!&\! -\frac{z_r \sin \theta}{||v(q_r)||} \!&\!  \cos \theta \!&\!  \frac{x_r \sin \theta}{||v(q_r)||}\\
\frac{z_r \sin \theta}{||v(q_r)||} \!&\!  \frac{y_r \sin \theta}{||v(q_r)||} \!&\! -\frac{x_r \sin \theta}{||v(q_r)||} \!&\!  \cos \theta
\end{array}
\!\!\right ]\!\!,
\end{flalign*}
$\theta\!=\! \cos^{-1} \frac{w_r}{||q_r||}$, $\phi \!=\! \cosh^{-1} ||q_r||$,  $\frac{\overline{q_r}q_i}{||q_r||||q_i||} \!=\! a\bfi \!+\! b\bfj \!+\! c \bfk$, and $\theta, \phi, a,b,c \in \bbR$. 
Alternatively, $\calM(q)$ can be factorized as $\calM(q) \!=\! \calM(u)\,\calM(h')$, where $\frac{q_i\overline{q_r}}{||q_i||||q_r||} \!=\! a'\bfi \!+\! b'\bfj \!+\! c' \bfk$, and $\calM(h')\!=\!$
\vspace{-\topsep}
\begin{equation*}
\resizebox{1.05\linewidth}{!}{
\begin{math}
\begin{aligned}
&\left [\!\!
\begin{array}{rrrr}
\cosh \phi \!&\! -a' \bfI\sinh\phi \!&\!- b' \bfI\sinh\phi \!&\! - c' \bfI\sinh\phi\\
a' \bfI\sinh\phi \!&\!  \cosh \phi \!&\!  c' \bfI\sinh\phi \!&\! -b' \bfI\sinh\phi\\
b' \bfI\sinh\phi \!&\! -c' \bfI\sinh\phi \!&\!  \cosh \phi \!&\!  a' \bfI\sinh\phi\\
c' \bfI\sinh\phi \!&\!  b' \bfI\sinh\phi \!&\! -a' \bfI\sinh\phi \!&\!  \cosh \phi
\end{array}
\!\!\right ]\!\!. %
\end{aligned}
\end{math}}
\end{equation*}
In addition, the determinants of $\calM(h)$, $\calM(h')$ and $\calM(u)$ are 1, and $\calM(h)$, $\calM(h')$ and $\calM(u)$ are orthogonal.
\end{customthm}
\begin{proof}
We utilize the ansatzes $u = \frac{q_r}{||q_r||}$, $h = ||q_r|| + \bfI \frac{\overline{q_r}q_i}{||q_r||||q_i||} ||q_i||$, and $h' = ||q_r|| +$ $ \bfI \frac{q_i\overline{q_r}}{||q_i||||q_r||} ||q_i||$ 
Note that $u$ is a unit quaternion ($q_r$ normalized), and $h$ 
is a biquaternion in which
$\frac{\overline{q_r}q_i}{||q_r||||q_i||}$
is a quaternion that is both pure (using Lemma~\ref{lem:pure})
and of unit norm (it is a product of two unit quaternions $\frac{\overline{q_r}}{||q_r||}\!=\!\frac{\overline{q_r}}{||\overline{q_r}||}$ and $\frac{q_i}{||q_i||}$). 
Similarly, 
$ h'$ is a biquaternion in which
$\frac{q_i\overline{q_r}}{||q_i||||q_r||}$ is a quaternion that is both pure and of unit norm.
(NB: if  $\alpha$ and $\beta$ are unit quaternions, then $||\alpha\beta||^2 = \overline{\alpha\beta}\alpha\beta = \overline{\beta}\overline{\alpha}\alpha\beta = \overline{\beta}\beta = 1$.) 

\vspace{-\topsep} 
\begin{align*}
uh =& \frac{q_r}{||q_r||} \left(||q_r|| + \bfI \frac{\overline{q_r}q_i}{||q_r||||q_i||} ||q_i||\right) \\
     =& q_r + \bfI \frac{q_r\overline{q_r}q_i}{||q_r||||q_r||} = q_r + \bfI \frac{||q_r||^2q_i}{||q_r||^2} = q
\end{align*}
Similarly, 
\vspace{-\topsep} 
\begin{align*}
h'u =&  \left(||q_r|| + \bfI \frac{q_i\overline{q_r}}{||q_i||||q_r||} ||q_i||\right)\frac{q_r}{||q_r||} \\
     =& q_r + \bfI \frac{q_i\overline{q_r}q_r}{||q_r||||q_r||} \\
     =& q_r + \bfI \frac{q_i||q_r||^2}{||q_r||^2} = q \;\; \mbox{(by associativity)}
\end{align*}
Using Equation~\ref{eqn:mmproduct},
we get the following factorizations: %
$\calM(q)\!=\!\calM(uh) \!=\! \calM(h)\calM(u)$, and $\calM(q)\!=\! \calM(h'u) \!=\! \calM(u)\calM(h')$. 

From Section~\ref{sec:background}, we know that the quaternion $u \!=\! \frac{q_r}{||q_r||}\!=\!\frac{w_r}{||q_r||}\!+\!\frac{x_r}{||q_r||}\bfi \!+\! \frac{y_r}{||q_r||}\bfj \!+\! \frac{z_r}{||q_r||}\bfk$ can be represented equivalently as $u \!=\! ||u|| (\cos \theta \!+\! \frac{v(u)}{||v(u)||} \sin \theta)$ 
$\!=\! \cos \theta \!+\! \frac{v(q_r)}{||v(q_r)||} \sin \theta$ (where $\theta \!=\! \cos^{-1} \frac{w_r}{||q_r||}$; $\theta \!\in\! \bbR$ because $w_r, ||q_r|| \!\in\! \bbR$). Expanding $v(q_r)$, we get 
$u = \cos \theta \!+\!\frac{x_r\sin\theta}{||v(q_r)||}\bfi \!+\! \frac{y_r\sin\theta}{||v(q_r)||}\bfj \!+\! \frac{z_r\sin\theta}{||v(q_r)||}\bfk$. Using the matrix representation given by Equation~\ref{eqn:matrixrep}, we get the form of $\calM(u)$ as stated in the theorem.

Since $\frac{\overline{q_r}q_i}{||q_r||||q_i||}$ is a pure unit quaternion, it can be represented as 
$a \bfi \!+\! b \bfj \!+\! c \bfk$ ($a,b,c \in \bbR$).
Thus, $h = ||q_r|| + \bfI (a \bfi \!+\! b \bfj \!+\! c \bfk) ||q_i||$ where $\frac{\overline{q_r}q_i}{||q_r||||q_i||} = a \bfi \!+\! b \bfj \!+\! c \bfk$. From Lemma~\ref{lem:pure}, $||q_r|| = \cosh \phi$ and $||q_i|| = \sinh \phi$. Thus 
$h = \cosh \phi \!+\!  (a\bfI\sinh\phi) \bfi \!+\! (b\bfI\sinh\phi) \bfj \!+\! (c\bfI\sinh\phi) \bfk$. Similarly, we can obtain $h' = \cosh \phi +  (a'\bfI\sinh\phi) \bfi + (b'\bfI\sinh\phi) \bfj + $ $ (c'\bfI\sinh\phi) \bfk$ and $\frac{q_i\overline{q_r}}{||q_i||||q_r||} \!=\! a'\bfi \!+\! b'\bfj \!+\! c'\bfk$.
Again, using the the matrix representation given by Equation~\ref{eqn:matrixrep}, we get the form of $\calM(h)$ and $\calM(h')$ as stated in the theorem.

Note that all elements in $\calM(h), \calM(h')$, and $\calM(u)$ are derived from $q$. Hence, for any $q$, we can construct $\calM(h)$, $\calM(h')$, and $\calM(u)$, thus proving that $\calM(q)$ can be factorized as $\calM(q) \!=\!\calM(h)\calM(u)$ and $\calM(q) =\calM(u)\calM(h')$.

Next, we show that since both $u$ and $q$ are unit biquaternions, $h$ and $h'$ is each a unit biquaternion.
$q = uh \Rightarrow \bar{q}q = \overline{uh}uh = \bar{h}\bar{u}uh = \bar{h}h = ||h||^2$. Likewise, $q = h'u \Rightarrow q\bar{q}= h'u\overline{h'u} = h'u\bar{u}\bar{h'} = h'\bar{h'} = ||h'||^2$. Since 
$q$ is a unit biquaternion, $\bar{q}q = q\bar{q} = 1 =$ $  ||h||^2 = ||h'||^2$. 

Using Theorem~\ref{thm:jafari_det}, we can obtain $\det \calM(h) \!=\! ||h||^4\!=\!1$, $\det \calM(h') \!=\! ||h'||^4 \!=\!1\!$, and $\det \calM(u) \!=\! ||u||^4 \!=\! 1$.

$\bar{h}h \!=\! 1 \Rightarrow \calM(h)\calM(\bar{h}) \!=\! \calM(h)\calM(h)^T\!=\! \calM(1) \!=\! \mathbb{I}$. 
Likewise, $h'\bar{h'} \!=\! 1 \Rightarrow \calM(\bar{h'})\calM(h') \!=\!$ $ \calM(h')^T\calM(h')\!=\! \calM(1) \!=\! \mathbb{I}$, $\bar{u}u \!=\! 1 \Rightarrow \calM(u)\calM(\bar{u})$ $ \!=\! \calM(u)\calM(u)^T \!=\! \calM(1) \!=\! \mathbb{I}$. 
Thus $\calM(h), \calM(h')$, and $\calM(u)$ are orthogonal.
\end{proof}

\section{Analysis and Visualization of BiQUE's Circular and Hyperbolic Rotations}
\label{sec:B}
\begin{figure}[t]
 \includegraphics[width=\linewidth]{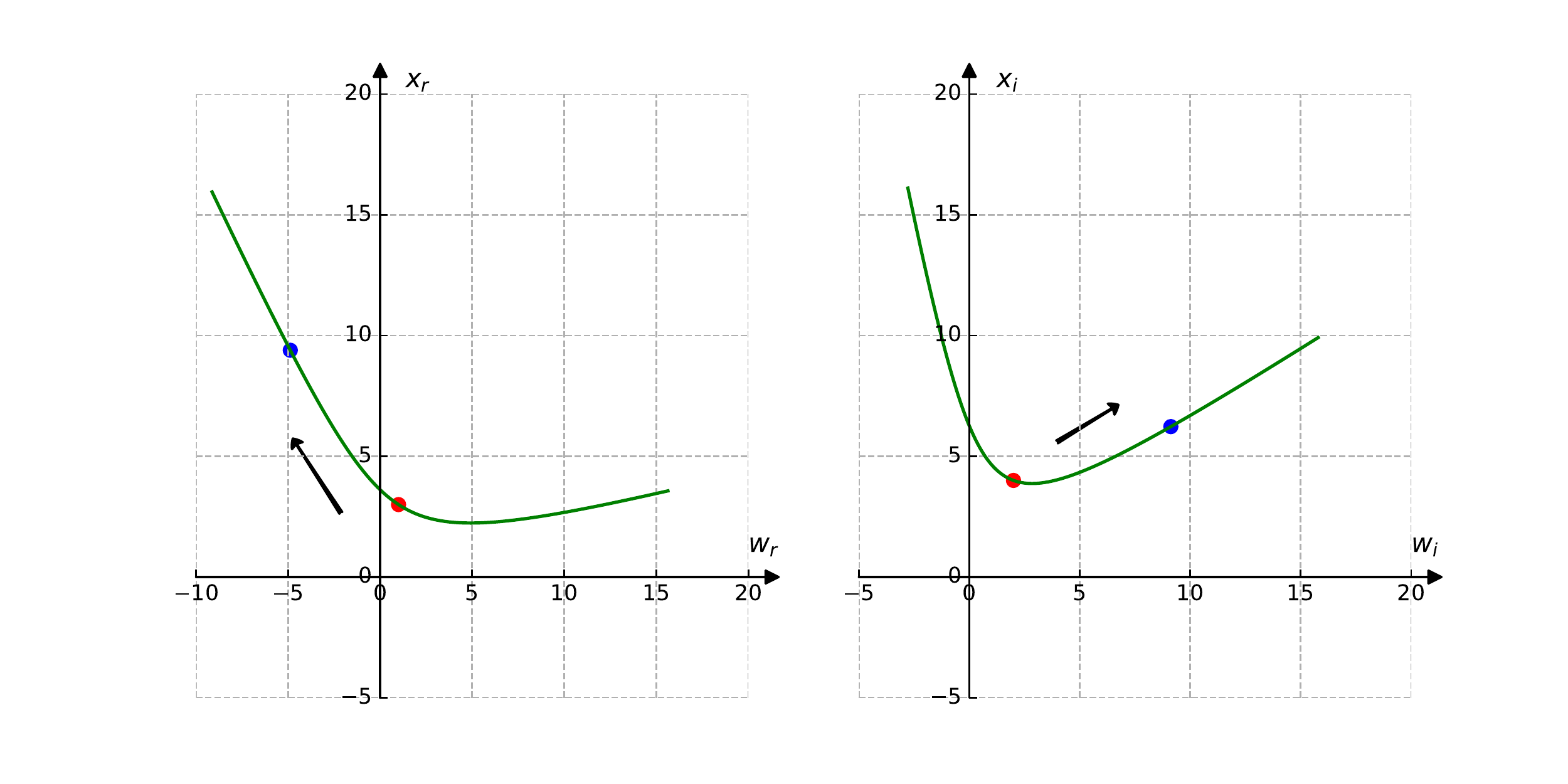}
 \caption{$M(h)$'s hyperbolic rotations in the real and imaginary parts respectively.} \label{fig:hyb}
\end{figure}

To analyze BiQUE's circular and hyperbolic rotations, we restrict ourselves to two dimensions. This means a biquaternion takes the form of $q = w + x \bfi$ where $w, x \in \bbC$. The unit quaternion $q_r$ in Theorem~\ref{thm:cirhyp} is thus $q_r = w_r + x_r \bfi$ where $w_r, x_r \in \bbR$, and $x_r/||v(q_r)|| = x_r/||x_r\bfi|| = 1$. The circular rotation matrix is thus
\begin{flalign*}
M(u) =\begin{bmatrix}
    \cos \theta & -\sin \theta \\
    \sin \theta & \cos \theta
    \end{bmatrix}
\end{flalign*}
Now, we multiply $M(u)$ with an arbitrary biquaternion $(w_r + w_i \bfI) + (x_r + x_i \bfI) \bfi$ to transform it.
\begin{equation*}
\resizebox{1\linewidth}{!}{
\begin{math}
\begin{aligned}
&\begin{bmatrix}
    \cos \theta & -\sin \theta \\
    \sin \theta & \cos \theta
    \end{bmatrix}
\begin{bmatrix}
w_r + w_i \bfI \\
x_r + x_i \bfI
\end{bmatrix} \\ 
&\!=\! \begin{bmatrix}
    \cos \theta & -\sin \theta \\
    \sin \theta & \cos \theta
    \end{bmatrix}
\begin{bmatrix}
w_r \\
x_r 
\end{bmatrix} 
\!+\! \bfI \begin{bmatrix}
    \cos \theta & -\sin \theta \\
    \sin \theta & \cos \theta
    \end{bmatrix}
    \begin{bmatrix}
w_i \\
x_i 
\end{bmatrix} \\
\end{aligned}
\end{math}}
\end{equation*}
We can see that the real parts $w_r, x_r$ and imaginary parts $w_i, x_i$ are transformed independently. Hence we can accomplish the same effect by rotating two quaternions ($w_r + x_r \bfi$ and $w_i + x_i \bfi$) independently, and this does not imbue biquaternions with added representational power beyond that of quaternions, which also have the rotation $M(u)$ matrix.

Now, we examine the effect of the hyperbolic rotation matrix $M(h)$. Since $a\bfi+b\bfj+c\bfk$ is a unit quaternion (as shown in the proof of Theorem~\ref{thm:cirhyp}), and we restrict ourselves to two dimensions, it must be that $a=1, b=0, c=0$. The hyperbolic rotation matrix is thus
\begin{flalign*}
M(h) =\begin{bmatrix}
    \cosh \phi & - \bfI \sinh \phi \\
    \bfI \sinh \phi & \cosh \phi
    \end{bmatrix}
\end{flalign*}
We multiply $M(h)$ with an arbitrary biquaternion $(w_r + w_i \bfI) + (x_r + x_i \bfI) \bfi$ to transform the latter.
\begin{equation*}
\resizebox{1\linewidth}{!}{
\begin{math}
\begin{aligned}
&\begin{bmatrix}
    \cosh \phi & - \bfI \sinh \phi \\
    \bfI \sinh \phi & \cosh \phi
    \end{bmatrix}
\begin{bmatrix}
w_r + w_i \bfI \\
x_r + x_i \bfI
\end{bmatrix} \\ 
&= \begin{bmatrix}
w_r & x_i \\
x_r & -w_i
\end{bmatrix}
\begin{bmatrix}
\cosh \phi \\
\sinh \phi
\end{bmatrix} 
+ \bfI \begin{bmatrix}
w_i & -x_r \\
x_i & w_r
\end{bmatrix}
\begin{bmatrix}
\cosh \phi \\
\sinh \phi
\end{bmatrix}
\end{aligned}
\end{math}}
\end{equation*}

Observe each term in the sum now involves both the real and imaginary parts ($w_r,x_r,w_i,x_i$) of the input biquaternion. This is unlike the case above for $M(u)$ in which the real and imaginary components are independent. Thus it is the hyperbolic rotation $M(h)$  that allows for the interaction between the real and imaginary components. To illustrate the hyperbolic rotation, we set $w_r = 1, w_i = 2, x_r = 3,  x_i = 4$, and change the value of $\phi$ continually from an initial value of 0. Note that when $\phi = 0$, the first term in the sum is the point $(w_r, x_r)$ and the second term is $(w_i, x_i)$. As $\phi$ changes, we can visualize the projection of that point. In Figure~\ref{fig:hyb}, the initial points in red are projected along the green lines. Clearly the green paths are hyperbolic. The blue point is an example of a projected point.

\section{Normalization of biquaternions}
\label{sec:biqnorm}

Given that $Q_r^\times = (w_r+w_i\mathbf{I}) + (x_r+x_i\mathbf{I})\mathbf{i} + (y_r+y_i\mathbf{I})\mathbf{j} + (z_r+z_i\mathbf{I})\mathbf{k}$, let $A = (w_r^2 + x_r^2 + y_r^2 + z_r^2)$ and $B = (w_i^2 + x_i^2 + y_i^2 + z_i^2)$, we define the real vector norm $\|Q_r^\times\|_v$ and biquaternion norm $\|Q_r^\times\|_b$ as follows:
\begin{align*}
\begin{aligned}
\|Q_r^\times\|^2_v &= A + B\\
\|Q_r^\times\|^2_b &= A - B + 2(w_rw_i+x_rx_i+y_ry_i+z_rz_i)\mathbf{I} \\
\end{aligned}
\end{align*}
Thus, we can obtain $Q_r^{\times\star}$ in section~\ref{sec:variants} with the standard normalization of real vectors: $Q_r^{\times\star} = \frac{Q_r^\times}{\|Q_r^\times\|_v}$.%
To make $Q_r^\times$ be a unit biquaternion, we have to make sure that $A - B = 1$ and $w_rw_i+x_rx_i+y_ry_i+z_rz_i = 0$. We first employ the Gram-Schmidt orthogonalization technique to guarantee that the imaginary coefficient is zero and then restrict $B = 1$. Alternatively, we represent $Q_r^\times$ as $Q_r^\times = q_1 + q_2 \mathbf{I}$, and conduct the following operations:
\begin{align*}
q_1' &= q_1 - \frac{<q_1, q_2>}{\|q_2\|^2}q_2\\
\widetilde{q_1} &= \frac{\sqrt{2}q_1'}{\|q_1'\|}, \ \ \widetilde{q_2} = \frac{q_2}{\|q_2\|}. \\
\end{align*}
Thus, we obtain the unit biquaternion $Q_r^{\times\triangleleft} = \widetilde{q_1} + \widetilde{q_2} \mathbf{I}$.

\section{Variance of the performance}
In Table~\ref{tab:avg1} and~\ref{tab:avg2}, we provide the averages and standard deviations for all metrics on the FB15k-237, WN18RR, YAGO3-10, CN-100K and ATOMIC datasets. The results are reported based on 10 runs with different random initializations. We see that the performance of our model BiQUE is quite stable across different random initializations,  and this supports the robustness of our method.

\section{Implementation Details}

For training, we adopt reciprocal learning~\cite{N3}, in which we add an inverse triple $(t,r^{-1},h)$ for each observed triple $(h,r,t)$ in the training data. For model optimization, we use Adagrad~\cite{Adagrad} as the optimizer, and employ grid search to find the best hyperparameters according to their performances on validation sets. The hyperparamters we search over includes embedding size (\{128, 256, 512, 1024\}), batch size (\{300, 500, 1000, 2000, 5000\}), learning rate (\{0.1, 0.01\}), and regularization parameters of Equation~\ref{eqn:objfunc} 
($\lambda$:\{0, 5e-3, 1e-2, 5e-2, 7e-2, 1e-1, 1.5e-1\};
$\lambda_1$, $\lambda_2$: \{0.5, 1.0, 1.5, 2.0\}). The parameters used in Table~\ref{tab:main} are shown in Table~\ref{tab:hyperpara}. We implement our BiQUE model in PyTorch, and run all experiments on NVIDIA Quadro RTX 8000 GPUs.

\begin{table}[h]
\centering
\scalebox{0.75}{
\begin{tabular}{lrrrrrrr}
\toprule
Datasets  & Epoch & Lr & Batch & $k$ & $\lambda$ & $\lambda_1$ & $\lambda_2$ \\
\midrule
WN18RR      & 200 & 1e-1 & 300  & 128 & 1.5e-1 & 2.0 & 0.5  \\
FB15K-237   & 300 & 1e-1 & 500  & 128 & 7e-2   & 2.0 & 0.5 \\
YAGO3-10    & 200 & 1e-1 & 1000 & 128 & 5e-3   & 2.0 & 0.5 \\
CN-100K     & 200 & 1e-1 & 5000 & 128 & 1e-1   & 2.0 & 0.5 \\
ATOMIC      & 200 & 1e-1 & 5000 & 128 & 5e-3   & 2.0 & 0.5 \\
\bottomrule
\end{tabular}}
\caption{Best hyperparameters for benchmarks. Lr is learning rate. $\lambda, \lambda_1, \lambda_2$ are Equation~\ref{eqn:objfunc}'s hyperparameters.}
\label{tab:hyperpara}
\end{table}

\begin{table*}[ht!]
\centering
\scalebox{0.8}{
\begin{tabular}{lrrrrrrrrrrrr}
\toprule
 &  \multicolumn{4}{c}{WN18RR}  &  \multicolumn{4}{c}{FB15K-237} & \multicolumn{4}{c}{YAGO3-10} \\
\cmidrule(lr){2-5} \cmidrule(lr){6-9} \cmidrule(lr){10-13}
Models  & MRR & H@1 & H@3 & H@10 & MRR & H@1 & H@3 & H@10 & MRR & H@1 & H@3 & H@10 \\
BiQUE & 0.502 & 0.457 & 0.518 & 0.589 & 0.363 & 0.267 & 0.401 & 0.554 & 0.578 & 0.504 & 0.623 & 0.711\\
(STDEV) & 0.001 & 0.001 & 0.001 & 0.001 & 0.001 & 0.001 & 0.001 & 0.001 & 0.002 & 0.002 & 0.002 & 0.001\\
\bottomrule
\end{tabular}}
\caption{The average results and their standard deviations for BiQUE on WN18RR, FB15k-237 and YAGO3-10 datasets.}
\label{tab:avg1}
\end{table*}

\begin{table*}[ht!]
\centering
\scalebox{0.8}{
\begin{tabular}{lrrrrrrrr}
\toprule
& \multicolumn{4}{c}{CN-100K} & \multicolumn{4}{c}{ATOMIC} \\
\cmidrule(lr){2-5} \cmidrule(lr){6-9}
Models  & MRR  & H@1 & H@3 & H@10 & MRR & H@1 & H@3 & H@10   \\
\midrule
BiQUE &  0.319  & 0.210  & 0.363 & 0.550 & 0.191 & 0.171 & 0.195 & 0.229\\
(STDEV) & 0.001  & 0.003 & 0.003 & 0.003 & 0.000 & 0.000 & 0.000 & 0.001\\
\bottomrule
\end{tabular}}
\caption{The average results and their standard deviations for BiQUE on CN-100K and ATOMIC datasets.}
\label{tab:avg2}
\end{table*}

\begin{table}[ht]
\end{table}

\end{document}